\newcommand*{\ARXIV}{}  

\ifdefined\SMC
    \documentclass[conference,letter]{IEEEtran}
    \IEEEoverridecommandlockouts
\fi
\ifdefined\ARXIV
    \documentclass{article}
    \usepackage[utf8]{inputenc}
    \usepackage{arxiv}
    \usepackage{moreverb,url}
    \usepackage{amssymb}
\fi
\usepackage{booktabs}
\usepackage{makecell}
\usepackage[T1]{fontenc}
\usepackage{amsmath,amssymb,amsfonts}
\usepackage{algorithm, algpseudocode}
\usepackage[longtable]{multirow}
\usepackage{amssymb}
\usepackage{comment}
\usepackage{subcaption}
\usepackage{graphicx}
\usepackage{textcomp}
\usepackage[dvipsnames]{xcolor}
\usepackage{todonotes}
\usepackage{tabularx}
\ifdefined\SMC
    \usepackage{hyperref} 
\fi
\ifdefined\ARXIV
    \usepackage[bookmarks=false]{hyperref}
\fi
\usepackage{makecell}
\usepackage[backend = biber, style=ieee, hyperref = true, url=false, isbn=false, doi=false]{biblatex}
\DeclareLanguageMapping{british}{british-apa}
\newcommand{\SI}{}

\let\eqref\cref
\ifdefined\SMC
    \newcolumntype{Y}{>{\centering\arraybackslash}X}
\fi
\ifdefined\ARXIV
    \newcolumntype{Y}{>{\centering\arraybackslash}X}
\fi

\newcommand*{\secref}[1]{Sec.~\ref{#1}}
\newcommand*{\tabref}[1]{Table~(\ref{#1})}
\newcommand*{\figref}[1]{Fig.~(\ref{#1})}
\newcommand*{\eqref}[1]{Eq.~(\ref{#1})}

\def\BibTeX{{\rm B\kern-.05em{\sc i\kern-.025em b}\kern-.08em
    T\kern-.1667em\lower.7ex\hbox{E}\kern-.125emX}}
    
\newcommand{\titleName}{Improving generalization of deep fault detection models in the presence of mislabeled data}
\newcommand{\abstr}{
 Mislabeled samples are ubiquitous in real-world datasets as rule-based or expert labeling is usually based on incorrect assumptions or subject to biased opinions. Neural networks can "memorize" these mislabeled samples and, as a result, exhibit poor generalization. 
 This poses a critical issue in fault detection applications, where not only the training but also the validation datasets are prone to contain mislabeled samples. 
In this work, we propose a novel two-step framework for robust training with label noise. In the first step, we identify outliers (including the mislabeled samples) based on the update in the hypothesis space. In the second step, we propose different approaches to modifying the training data based on the identified outliers and a data augmentation technique.
Contrary to previous approaches, we aim at finding a robust solution that is suitable for real-world applications, such as fault detection, where no clean, "noise-free" validation dataset is available. Under an approximate assumption about the upper limit of the label noise, we significantly improve the generalization ability of the model trained under massive label noise. }

\newcommand{\keywo}{fault detection, label noise, generalization, deep neural networks, \textit{mixup}}

\newcommand{\aknow}{This research resulted from the "Integrated intelligent railway wheel condition prediction" (INTERACT) project, supported by the ETH Mobility Initiative.}

\begin{document}

\ifdefined\SMC
    \title{\titleName\\
    
    \thanks{\aknow}
    }
    
    \author{
    \IEEEauthorblockN{1\textsuperscript{st} Katharina Rombach}
    \IEEEauthorblockA{\textit{
    \small{Chair of Intelligent Maintenance Systems}} \\
    \textit{ETH Zurich}\\
    Zurich, Switzerland \\
    rombachk@ethz.ch}
    \and
    \IEEEauthorblockN{2\textsuperscript{nd} Gabriel Michau}
    \IEEEauthorblockA{\textit{\small{Chair of Intelligent Maintenance Systems}} \\
    \textit{ETH Zurich}\\
    Zurich, Switzerland \\
    gmichau@ethz.ch}
    \and
    \IEEEauthorblockN{3\textsuperscript{rd} Olga Fink,~\IEEEmembership{Member,~IEEE}}
    \IEEEauthorblockA{\textit{\small{Chair of Intelligent Maintenance Systems}} \\
    \textit{ETH Zurich}\\
    Zurich, Switzerland\\
    ofink@ethz.ch}
    }
    
    \maketitle

    \begin{abstract}%
    \abstr
    \end{abstract}
    
\fi
\ifdefined\ARXIV
\title{\titleName}
    \author{%
        Katharina Rombach\\
        \textit{\small{Chair of Intelligent}} \\\textit{\small{Maintenance Systems}}\\
        ETH Z\"urich,\\
        Z\"urich, Switzerland\\
        \And 
        Gabriel Michau\\
        \textit{\small{Chair of Intelligent}} \\\textit{\small{Maintenance Systems}}\\
        ETH Z\"urich,\\
        Z\"urich, Switzerland\\
        \And 
        Olga Fink\\
        \textit{\small{Chair of Intelligent}} \\\textit{\small{Maintenance Systems}}\\
        ETH Z\"urich,\\
        Z\"urich, Switzerland}
        \subtitle{Preprint}
        \date{\today}
        \maketitle
        \begin{abstract}
        \abstr
        \end{abstract}
\fi

\section{Introduction}
\label{sec:Introduction}

Many real-world datasets exhibit label noise \cite{krishna2016embracing}. In practical applications of fault detection, labels for distinguishing between healthy and faulty conditions are often generated by predefined rules or else based on assumptions. For example, a system is considered to be healthy within a defined period of time after a performed maintenance action. However, this assumption does not always hold, which results in mislabeled samples in both the training and validation  datasets.
While deep neural networks (NN) have been applied successfully in the field of Prognostics and Health Management (PHM) \cite{abdeljaber2017real,krummenacher2017wheel}, their performance is heavily impacted if trained on a dataset with label noise. As universal approximators, NNs are capable 
of fitting to any labels \cite{zhang2016understanding}. This ability is referred to as "memorization" \cite{arpit2017closer} and leads to poor generalization of the resulting models. In the absence of a clean validation dataset, this lack of generalization cannot be detected since the model might exhibit good performance on a validation dataset that is impacted by the same label bias as the training dataset. However, the model may not have learned the true relationship between input and output.  This is especially problematic in the context of fault detection in industrial assets, where faults are safety-critical.    

In this work, we tackle the challenge of training a robust model despite the presence of label noise and without exact knowledge about the label noise or a clean dataset with which to perform hyperparameter (HP) tuning. We propose a two-step framework that first identifies outliers based on the samples' consistency with the hypothesis update and second, modifies the training dataset based on the identified outlier samples. An adaptation of the data augmentation technique called \textit{mixup} \cite{zhang2017mixup} is introduced for the data modification. We aim at providing universally applicable recommendations for learning under label noise.

To the best of our knowledge, this is the first study to tackle the inability to tune certain HP if no reliable ground truth information is available. Our proposed solution relies only on a rough assumption regarding the level of label noise. Ultimately, we significantly improve the generalization ability of the trained models under massive label noise on an image dataset and a time series dataset for fault detection. 

After reviewing the existing relevant literature in \secref{sec:RelatedWork}, the proposed framework is introduced in \secref{sec:meth}.
The methodology is evaluated on the experimental setup as defined in \secref{sec:experiment} and the results are shown in \secref{sec:results}. Based on the discussion in \secref{sec:discussion}, final recommendations about training deep models with label noise are given in \secref{sec:conclusion}.

\section{Related Work}
\label{sec:RelatedWork}
Robust learning on noisy datasets has attracted increasing attention - especially in the context of deep learning. In the literature review, we aim at giving a brief overview of different approaches and their limitations with respect to the scenario relevant for fault detection. We elaborate in more detail only closely related methods. For a detailed survey on classification under label noise, the reader is referred to \cite{frenay2013classification}. 

\textbf{Direct approaches} aim at detecting mislabeled samples explicitly. E.g.\ all samples are ranked by their probability of being assigned to the original label (based on the current model's prediction) and a fraction $\alpha$ of this ranked list is then presented to experts for relabeling \cite{muller2019identifying}. Hence, this approach i.a.\ relies on human intervention. 

Other approaches based on the model's prediction aim at automatic relabeling \cite{tanaka2018joint, reed2014training}. However, these tend to favor trivial solutions to the classification task.
To counteract this, they require prior knowledge, such as the prior distribution over all classes \cite{tanaka2018joint}. 
In addition to the model's prediction, logits \cite{pleiss2020identifying} and the training loss \cite{shen2018learning} were also considered to identify mislabeled samples. 

As an alternative to relabeling, several researchers have proposed altering the current model's prediction to match the label noise as in \cite{vahdat2017toward, patrini2017making, sukhbaatar2014training}. 
Yet, some approaches  presuppose e.g.\ the ground truth noise model \cite{vahdat2017toward, patrini2017making}.  

Others aim to learn this model but do not carefully focus on the memorization ability of DNNs \cite{tanaka2018joint}. We argue that a clean validation dataset is required to tune crucial HP such as when to start updating the noise model Q \cite{sukhbaatar2014training}. 

Learning to reweight samples of a noisy dataset has been proposed in the field of meta-learning \cite{jiang2017mentornet, ren2018learning}.
 For example, a meta-gradient step was proposed in \cite{ren2018learning} to reweight samples by evaluating the gradient directions based on a noise-free  dataset before the network is updated. These methodologies usually rely on a small, noise-free validation dataset \cite{jiang2017mentornet, ren2018learning} and are therefore not applicable in our setting.

\textbf{Indirect approaches} deal with mislabeled data only implicitly. They aim at robust optimization in general, resulting in good generalization of the model despite the presence of mislabeled samples. Modified loss functions have been proposed, including generalized cross entropy \cite{zhang2018generalized} and an information-theoretic loss function \cite{xu2019l_dmi}. Also, different regularization techniques have been shown to yield good generalization \cite{hu2020simple, arpit2017closer}. Each of the proposed approaches comes with a set of specific HP.
We argue that tuning these optimally requires a clean validation dataset, which is not available in the scenario considered here.

Vicinal Risk Minimization (VRM)  \cite{chapelle2001vicinal} has been proposed as an alternative to Empirical Risk Minimization (ERM) \cite{vapnik1998statistical}, and not only in the context of label noise. 
It relies on the assumption that the true density function of the data is smooth in the vicinity of any data point and therefore opts to represent it by a vicinity distribution instead of the empirical one as in ERM  \cite{vapnik2013nature}. Recently, VRM has been shown to stabilize the training of NNs on noisy data with \textit{mixup} \cite{zhang2017mixup}. The data is augmented by drawing samples from a generic vicinal distribution, resulting in convex combinations of the datapoints and their respective labels. Thus, linear behaviour between the classes is favored, which has been shown to prevent the model from overfitting to individual mislabeled samples.

In this work, we tackle the problem in a more general context compared to previous works, by relaxing two strong assumptions that do not hold for many practical applications: 1) we rely neither on a clean dataset for tuning the methodology nor 2) on the exact knowledge of the label noise. We only assume a rough upper estimation of the noise level. The proposed end-to-end approach does not require any human intervention. It combines elements of both direct approaches (outlier detection (OD)) and indirect approaches since the detected outliers are used to adapt the model training. Since \textit{mixup} has shown good performance in other contexts and introduces only one additional HP that can be related to the noise level estimation, we also evaluate its performance and compare it to the proposed approaches.

\section{Methodology}
\label{sec:meth}

\subsection{Problem Formulation}
\label{sec:notation}
In a classification task, we aim at finding a function $h$ that captures the true relationship between a variable $X$ and a label $Y$, which follow the joint distribution $P(X,Y)$. In reality, only a finite number of samples of this joint distribution are available - a finite dataset $\mathcal{D}$. The set of functions $h \in \mathcal{A}(\mathcal{D})$ that can be reached depends i.a.\ on the provided dataset $\mathcal{D}$ \cite{arpit2017closer}. Given a training dataset $\mathcal{D}^{'}$ with unknown label noise, we aim at finding a deep model $h$ that performs well on the underlying true but unknown data distribution $P(X,Y)$. Since no ground truth  is available, all HP can only be tuned on $\mathcal{D}^{'}$.

\subsection{Proposed Framework}
\label{sec:framework}

\begin{figure*}
\centering
\ifdefined\ARXIV
\includegraphics[width=0.8\linewidth]{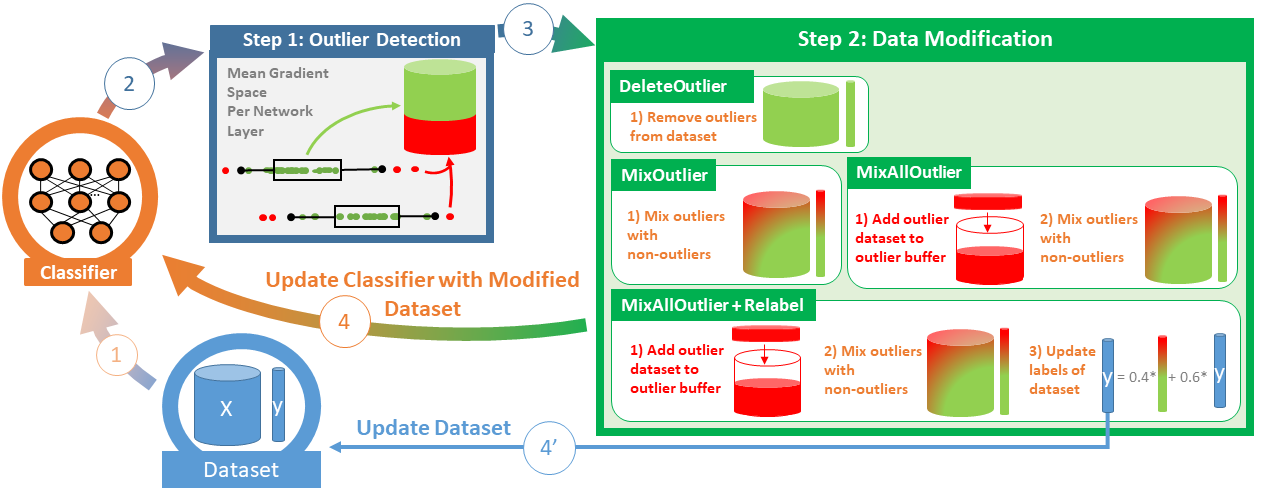}
\fi
\ifdefined\SMC
\includegraphics[width=0.68\linewidth]{Plots/label_noise_paper_graphics.png}
\fi
\caption{Process Flow.}
\label{fig:flow}
\end{figure*}

We introduce a novel two-step framework for robust training with label noise (see \figref{fig:flow}). In a first step, we identify a set of outliers including mislabeled samples (see \secref{sec:OD} and \textbf{\textcolor{BlueViolet}{Step 1}} in \figref{fig:flow}). The novel algorithm aims at \textbf{early detection} based on the gradient update in the hypothesis space, i.e.\  mislabeled samples are identified before they have been considered for the model update in order to prevent the NN from overfitting to these samples. 

In a second step, the training data is modified based on the previously identified outliers. A new adaptation of the data augmentation technique called \textit{mixup} is proposed. 
In \eqref{eq:mixup},
\begin{equation} \label{eq:mixup}
\begin{split}
\tilde{x} & = \lambda x_i + (1-\lambda)x_j \\
\tilde{y} & = \lambda y_i + (1-\lambda)y_j
\end{split}
\end{equation}
the original \textit{mixup} augmentation of $(x_{i},y_{i})$ is defined as proposed in \cite{zhang2017mixup},  where $\lambda \sim Beta(\alpha,\alpha)$ and the sample-label pairs $(x_{j},y_{j})$ are randomly chosen. The novel adaptation of the data augmentation is explained in \secref{sec:OD}. Multiple data modification techniques, including the automatic relabeling of the training dataset, are proposed in \secref{sec:meth_combine} (see\textbf{ \textcolor{Green}{Step 2}} in \figref{fig:flow}). %

\subsection{Assumptions}
\label{sec:assumptions}
We aim at developing a universal approach for the scenario in which label noise is suspected but no ground truth information is available. In such situations, necessity compels us to make only rough assumptions. Hence, we make the following rough assumptions regarding the label noise:
\begin{enumerate}
    \item the number of mislabeled samples is less than 50\% of the entire dataset
    \item we use a rough estimate of the upper limit of label noise (little, medium, massive) to set a maximum threshold of possibly detected outliers and to set the HP
\end{enumerate}

We argue that this is a very loose assumption compared to those required in previous studies where, e.g.\, the exact label noise must be known \cite{tanaka2018joint}. 
We also evaluate extreme scenarios in which the estimated upper noise limit is 10-20\% above the actual noise ratio. Furthermore, we assess the sensitivity of the proposed framework to these assumptions. To evaluate potential limitations, the scenarios are evaluated for cases in which the assumptions concerning the noise level are wrong (see \secref{sec:discussion}).

\subsection{Outlier Detection}
\label{sec:OD}
Outliers are defined as samples that are inconsistent with the update in the hypothesis space, i.e.\ samples with gradients surpassing certain thresholds as defined in Algorithm~\ref{algo:OutlierDetection}. The gradient $\partial_{i}^k$ in the parameter space for all samples $i$ and nodes $k$ is calculated (see line~\ref{lst:line:partial} in Algorithm~\ref{algo:OutlierDetection}), representing the gradient distribution in the parameter space. The thresholds are set based on the confidence interval of the gradient distribution: $wh_{lower}^l$ and $wh_{upper}^l$ are set to be below the 25th percentile and above the 75th percentile by a factor of 1.5 of the Interquartile Range (IQR) (see lines~\ref{lst:line:wh_start} - \ref{lst:line:wh_stop}  in Algorithm~\ref{algo:OutlierDetection}). Furthermore, to enable convergence, a minimal threshold is set for the IQR. The gradients are calculated with respect to the cross-entropy loss as it emphasizes difficult samples \cite{zhang2018generalized}. For all proposed approaches, hard labels  (the maximum-a-posterior (MAP) estimates) are used for the gradient calculation.
Furthermore, we use assumptions about the upper limit of the label noise to set a maximum number of outliers that can be detected. This limit is defined in \tabref{table:MixUP_ParaMeter}.
All samples that are not detected as outliers are considered consistent in updating the hypothesis. 

\textbf{Computational Complexity:} Since the parameter space of NNs can be very high-dimensional, the proposed outlier selection algorithm is computationally expensive.  Thus, we propose to represent the gradient distribution over all samples in a compact way by
only considering the mean gradient per layer $l \in L$ of the NN, defined as

\begin{equation}
\partial_{i,mean}^l =\frac{1}{K} \sum_{k \in K} \bigg(\frac{\partial f_{i}(w)}{\partial w_{k,l}}\bigg),
\label{eq:layer_gradient}
\end{equation}
where $f$ is the loss function, $k \in K$ the nodes in the respective layers $l \in L$  and $w$ the weights of the NN. 

Outliers are defined as samples whose mean gradient surpasses the defined thresholds in any of the layers.
Furthermore, we only consider the weights of the NN and neglect the biases.  
In this research, the outlier selection is performed for each class individually. It is important to note that this is not necessary but rather a design choice. The per-class detection enables the approach to be applied to imbalanced datasets as well. This makes the proposed approach more universally applicable. 

\ifdefined\ARXIV
    {\centering
    \begin{minipage}{.55\linewidth}
    \begin{algorithm}[H]
\fi
\ifdefined\SMC
    \begin{algorithm}
\fi

\begin{algorithmic}[1]
\Procedure {DetectOutlier}{$\mathcal{D}^{'}$, $h$}
\State $C$:  \#  classes $\in \mathcal{D}^{'}$
\State $L$:  \#  layers $\in h$
\ForAll {$c \in C$}
\ForAll {$l \in L$}
\ForAll {$(x_i,y_i) \in c$}
\State $\partial_{i,mean}^l =\frac{1}{K} \sum_{k \in K} \bigg(\frac{\partial f_{i}(w)}{\partial w_{k,l}}\bigg)$ \label{lst:line:partial}
\EndFor
\State $p_{25}^{l} = \mathrm{Percentile}_{25}(\{\partial_{i,mean}^l\}_{i\in c}) $ \label{lst:line:wh_start}
\State $p_{75}^{l} = \mathrm{Percentile}_{75}(\{\partial_{i,mean}^l\}_{i\in c}) $
\State $IQR^{l} = p_{75}^{l} - p_{25}^{l}$
\State $wh_{c,low}^{l} = p_{25}^{l} - 1.5*IQR^{l}$
\State $wh_{c,up}^{l} = p_{75}^{l} + 1.5*IQR^{l}$ \label{lst:line:wh_stop}
\EndFor
\EndFor
\State $\mathcal{O} = \{ i  \mid i \in c; \exists l \in L; \partial_{i,mean}^l \notin [wh_{c,low}^l, wh_{c,up}^l] \} $
\State $\mathcal{D}^{*} = \mathcal{D}^{'} \setminus \mathcal{O}$  
\State \textbf{return} $\mathcal{O}, \mathcal{D}^{*}$
\EndProcedure
\end{algorithmic}
\caption{Outlier Detection}
\label{algo:OutlierDetection}

\ifdefined\ARXIV
        \end{algorithm}
    \end{minipage}
    \par
    }
\fi
\ifdefined\SMC
    \end{algorithm}
\fi

\subsection{Data Modification}
\label{sec:meth_combine}
We propose different approaches to stabilizing the training through data modification using the detected set of outliers.  These  range from ERM on the non-outliers to VRM on the complete dataset. All approaches are visualized in \textcolor{Green}{\textbf{Step 2}} of \figref{fig:flow}.

More concretely, we propose to enforce different degrees of data augmentation, i.e.\ of linear interpolation between outliers and non-outliers,  by using different values of  $\alpha$ in \eqref{eq:mixup}. Furthermore, all samples are only mixed with non-outliers ($x_j$ in \eqref{eq:mixup}). We refer to this as \textit{Adapted MixUp}. If no outliers are detected, we optimize based on the empirical distribution, i.e. $\alpha=0$.  Thereby, memorization of outliers (incl. mislabeled samples) is prevented while, simultaneously, the model is still able to learn  nonlinear relationships based on the dataset that is consitent with the hypothesis.
The different approaches for the data augmentation step are introduced below: 
\ifdefined\SMC
    \\ %
    \textbf{DeleteOutlier: } 
\fi
\ifdefined\ARXIV
    \begin{description}
    \item[\textbf{DeleteOutlier: }]
\fi
  The identified outliers are removed from the training set for the next model update, i.e. for the next ERM optimization step. Yet the dataset for the OD remains unaltered, i.e.\ the OD is always performed on the original dataset. 
\ifdefined\SMC
    \\ %
     \textbf{MixOutlier: }
\fi
\ifdefined\ARXIV
    \item[\textbf{MixOutlier: }]
\fi
  All samples are augmented with \textit{Adapted MixUp}. A higher value of $\alpha$ is used on the currently detected set of outliers and a lower one for the current set of non-outliers. Again, the dataset for the OD is left unaltered.
\ifdefined\SMC
    \\ %
     \textbf{MixAllOutlier: }
\fi
\ifdefined\ARXIV
    \item[\textbf{MixAllOutlier: }]
\fi

 This approach is equivalent to \textbf{MixOutlier} except that samples that were detected as an outlier in any of the iterations are treated as outliers in each subsequent augmentation step. 
\ifdefined\SMC
    \\ %
    \textbf{MixAllOutlier + Relabel: }
\fi
\ifdefined\ARXIV
    \item[\textbf{MixAllOutlier + Relabel: }]
\fi

 This is based on \textbf{MixAllOutlier}. However, the labels in the dataset are permanently altered by building convex combinations with a factor of 0.6 from the label and the current prediction of the model - similar to \citeauthor{reed2014training} \cite{reed2014training}.
\ifdefined\ARXIV
    \end{description}
\fi

\section{Experimental Setup}
\label{sec:experiment}
The proposed two-step framework is evaluated based on two different datasets with different characteristics. We aim at evaluating the following properties: (1) Exp.1: different training dynamics on two datasets given symmetric label noise; (2) Exp.2: feasibility of \textit{Adapted MixUp}; (3) Exp.3: performance of the proposed OD approach and (4) Exp.4: the generalization capabilities of the resulting deep models evaluated on a clean test dataset. All experiments are repeated 5 times and the mean and standard deviation are reported. Symmetric label noise is added to the inherently clean training datasets as described in \cite{chen2019understanding, jiang2017mentornet, ma2018dimensionality}. 

For evaluation purposes only, the test dataset is not corrupted by label noise. 
Our evaluation focuses on binary classification tasks as this is a relevant setup for fault detection. However, the proposed framework is also applicable to multi-class classification tasks.
 We compare the proposed framework, comprising the OD combined with the different variants of data augmentation, to the two baseline methods ERM and \textit{mixup}. 

\subsection{Datasets}
\label{sec:datasets}
We evaluate the approaches on the MNIST dataset  \cite{lecun1990handwritten} containing images of handwritten digits from 0 to 9. We reformulate it as a binary classification task for demonstration purposes - i.e.\ digits 0-3 are grouped into class 0 (30596 training samples, 5139 test samples) and digits 4-9 are grouped into class 1 (29404 training samples, 4862 test samples) - to simulate defect detection in PHM where fault types, as well as the healthy states, can have multiple patterns (which can be regarded as different operating conditions). The dataset contains 60000 training samples and 10000 test samples in total and has previously been used as a benchmark dataset for anomaly detection in PHM applications \cite{ducoffeanomaly}. The images are normalized.

Furthermore, we apply the proposed framework to a simulated 
time series dataset - the Building Defect Detection dataset (BDD dataset) - to detect faults in buildings \cite{granderson2020building}. We conducted the experiments on the multi-zone variable air volume AHU dataset (MZVAV-2-2). 
The dataset contains measurements in a healthy state as well as measurements from three different fault patterns (leaking valve of heating coil, stuck valve of cooling coil, and stuck outdoor air damper). The sensor readings are recorded once per minute over 26 days.
Half-hour time windows are selected for the classification as this is sufficient to distinguish healthy from faulty conditions. In total, this resulted in 1247 sample-label pairs, of which 623 are considered healthy and 624 are considered faulty. 20\% of the data is randomly selected for the test dataset. In total, we consider 15 sensor readings,  i.e.\ all besides one set point and one control signal (AHU: Supply Air Temperature Set Point, AHU: Exhaust Air Damper Control Signal). 
Therefore, one sample consists of 450 measurements (30 minutes x 15 sensors). The measurements for each sensor are standardized.

\subsection{Hyperparameter Settings}
\label{sec:Hyperparameter}

As we assume that no clean, reliable validation dataset is available for HP tuning, we can only rely on metrics based on the training dataset and on the aforementioned assumptions. All parameters regarding the model and the standard optimization algorithm were chosen such that a training accuracy of at least 75\% is achieved for all label noise ratios for each dataset under \textit{ERM}. 

The model used for MNIST dataset is a four-layer fully connected NN with 128, 32, 10 nodes activated by ReLU, and 2 nodes activated by the sigmoid function. It is updated with Adam \cite{kingma2014adam} (initial learning rate of 0.001). The batch size is set to 64. 
The model used for the two BDD datasets is a five-layer fully connected NN with 256, 128, 64, 16 nodes activated by ReLU, and 2 nodes activated by the sigmoid function.  The optimizer Adam \cite{kingma2014adam} is used (initial learning rate of 0.0001) and the batch size is set to 16. Both models are trained by minimizing the cross-entropy loss as well as the entropy calculated based on the model's prediction as an additional regularization loss \cite{tanaka2018joint}. 

The minimal IQR for the OD is set heuristically to 0.0001 and is kept constant over all experiments - see \secref{sec:OD}. For \textit{mixup}, the default values for $\alpha$  proposed in the original paper are used \cite{zhang2017mixup} based on a basic assumption about the label noise as defined in  \tabref{table:MixUP_ParaMeter}. Unless stated otherwise, for our proposed approaches, we set $\alpha=0.4$ for the set of non-outliers $\mathcal{D}^{*}$ (as it corresponds to the assumption of \textit{Little Noise}) and $\alpha=32$ for the outlier dataset (corresponding to the \textit{Massive Noise} setting). As mentioned above, if no outliers are detected at all during the training process, we revert to \textit{ERM} by using a value of  $\alpha=0$.

\begin{table}
\caption{Setting of $\alpha$ under different noise assumptions based on proposed values in \cite{zhang2017mixup}.}
\label{table:MixUP_ParaMeter}
    \noindent\begin{center}
        \begin{tabularx}{\linewidth}{l| *{3}{Y}} 
        \toprule 
        \thead{\textbf{Assumption}} & \thead{\textbf{Noise} \\ \textbf{Ratio}} & \thead{\textbf{Upper} \\ \textbf{Threshold}} & {\thead{$\mathbf{\alpha}$}}\\ 
        \midrule 
        \midrule 
        \textit{Little Noise} & $0 - <10\%$ & $10\%$ & 0.4\\
        \textit{Medium Noise}  & $10 - <30\%$   & $30\%$ &  8\\
        \textit{Massive Noise}  & $30 - <50\%$  & $50\%$ & 32\\
        \bottomrule 
        \end{tabularx}
    \end{center}
\end{table}

\section{Results}
\label{sec:results}
\subsection{Experiment 1 - Training Dynamics on Mislabeled Data}
\begin{figure}
\centering
\begin{subfigure}{0.7\columnwidth}
    \includegraphics[width=.85\linewidth]{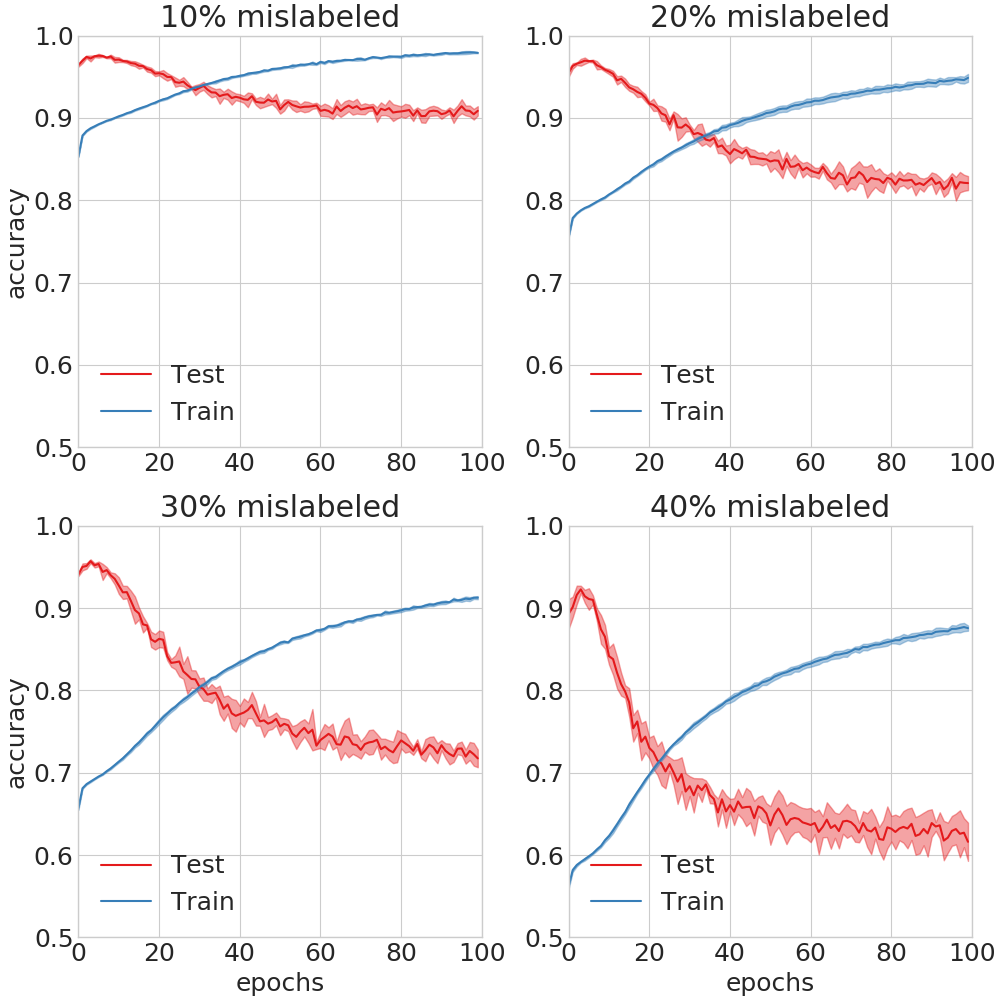}
    \caption{MNIST dataset.}
    \label{fig:MNIST_ERM}
\end{subfigure}
\begin{subfigure}{0.7\columnwidth}
    \includegraphics[width=.85\linewidth]{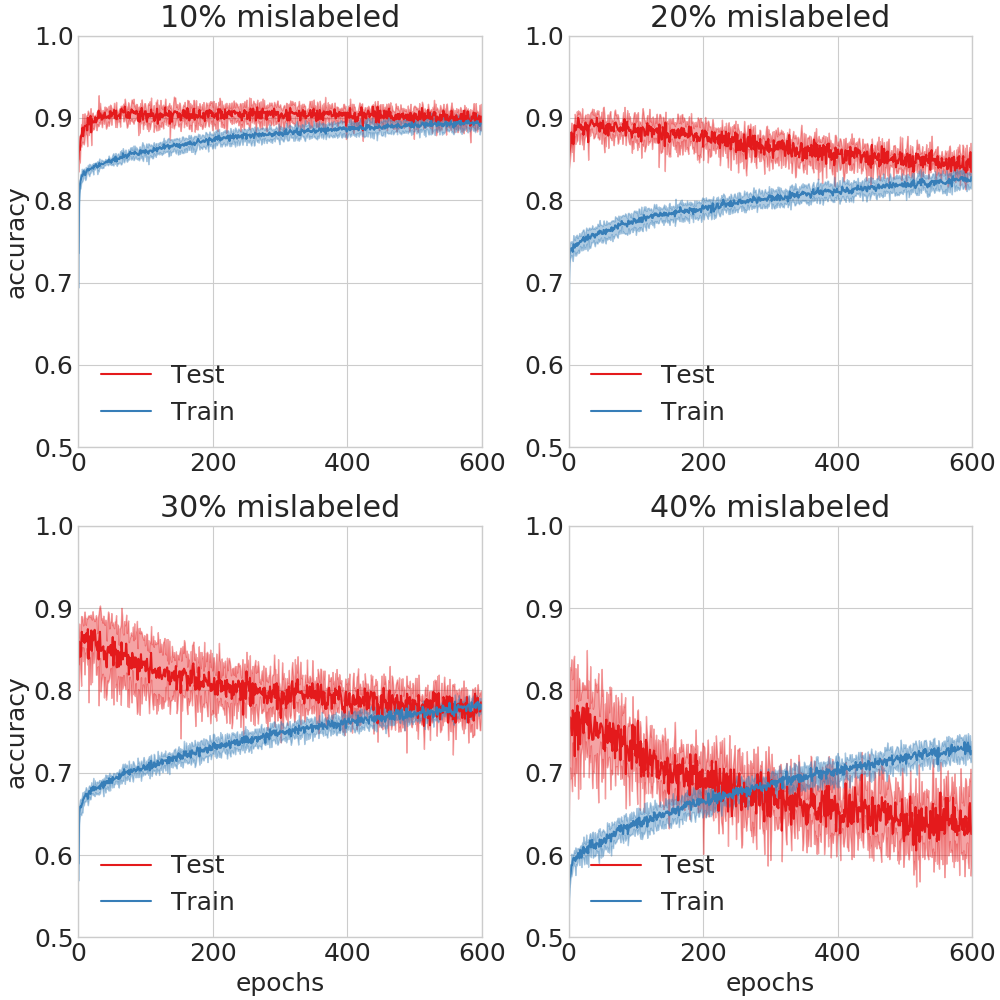}
    \caption{Building Defect Detection dataset.}
    \label{fig:BUILD_ERM}
\end{subfigure}
\caption{Training and Test Accuracy.}
\end{figure}

As preliminary results, we demonstrate the overfitting behaviour leading to poor generalization as described in the introduction. For this purpose, we train a model with ERM on a noisy dataset and plot the accuracy on the mislabeled training dataset and the test accuracy on a "clean" test dataset. 
In \figref{fig:MNIST_ERM} and \figref{fig:BUILD_ERM}, results  with different noise ratios are plotted. 
This demonstrates how the models overfit to noisy labels. 
 
\subsection{Experiment 2 - Adapted MixUp}
\begin{table}
\caption{Final Accuracy on Clean Test Dataset with Ground Truth Information.}
\label{table:PART_MIX}
\begin{center}
\begin{tabularx}{\columnwidth}{l| l| *{4}{Y}} 
\toprule
&& \multicolumn{4}{c}{\thead{\textbf{Label Noise}}} \\ 
 \thead{\textbf{Dataset}} & \thead{\textbf{Method}}  &  \thead{\textbf{10\%}} &  \thead{\textbf{20\%}} &  \thead{\textbf{30\%}} &  \thead{\textbf{40\%}}\\
\midrule
\midrule
\multirow{2}{*}{MNIST} &\textit{mixup}       & $\SI{97 \pm 0}\%$ & $\SI{95 \pm 0}\%$ & $\SI{90 \pm 1}\%$ & $\SI{80 \pm 1}\%$    \\
&\textit{MixOutlier}  & $\mathbf{\SI{98 \pm 0}\%}$ & $\mathbf{\SI{98 \pm 0}\%}$   & $\mathbf{\SI{98 \pm 0}\%}$  & $\mathbf{\SI{98 \pm 0}\%}$  \\
\midrule
\multirow{2}{*}{BDD} &\textit{mixup}       & $\mathbf{\SI{93 \pm 0}\%}$ & $\SI{90 \pm 1}\%$ & $\SI{87 \pm 1}\%$ & $\SI{78 \pm 1}\%$    \\
&\textit{MixOutlier}  & $\mathbf{\SI{93 \pm 1}\%}$  & $\mathbf{\SI{93 \pm 0}\%}$  & $\mathbf{\SI{93 \pm 2}\%}$  & $\mathbf{\SI{94 \pm 0}\%}$  \\
\bottomrule
\end{tabularx}
\end{center}
\end{table}

As a preliminary exploration, we evaluate the feasibility of enforcing different degrees of interpolation as described in \secref{sec:meth_combine} to demonstrate the rationale behind the proposed methodology. Therefore, we assume that we know the ground truth labels of the noisy training set and can thus identify the mislabeled samples. The results are compared to the original \textit{mixup} augmentation 
\cite{zhang2017mixup}. While $\alpha$ for the original \textit{mixup} is set as stated in  \secref{sec:Hyperparameter}, for the \textit{Adapted MixUp} a fixed value of  $\alpha = 32$ is set for the mislabeled samples and a value of  $\alpha = 0$ is set for the non-outliers as it is known to be noise-free given the ground truth information. 

 The results for $10\%\! - \!40\%$ label noise are shown in \tabref{table:PART_MIX}. For $0\%$ label noise, the approach is equivalent to \textit{ERM} under the assumption that ground truth information is available. Therefore, this evaluation is not listed in the table.
The \textit{Adapted MixUp} augmentation, given ground truth information, leads to  better generalization capabilities compared to default \textit{mixup} - especially under massive noise. The final performance of \textit{Adapted MixUp} is independent of the label noise ratio for both datasets.

\subsection{Experiment 3 - Outlier Detection}
\begin{figure*}
\centering
\begin{subfigure}{0.7\textwidth}
   \includegraphics[width=1\linewidth]{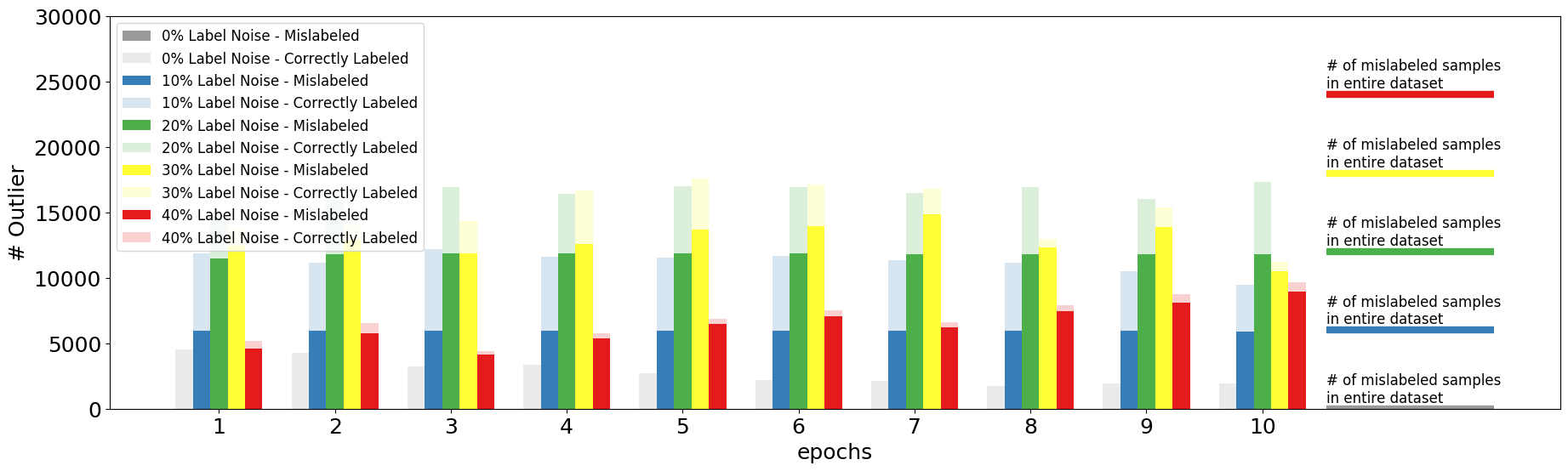}
   \caption{MNIST}
   \label{fig:OD_MNIST} 
\end{subfigure}

\begin{subfigure}{0.7\textwidth}
   \includegraphics[width=1\linewidth]{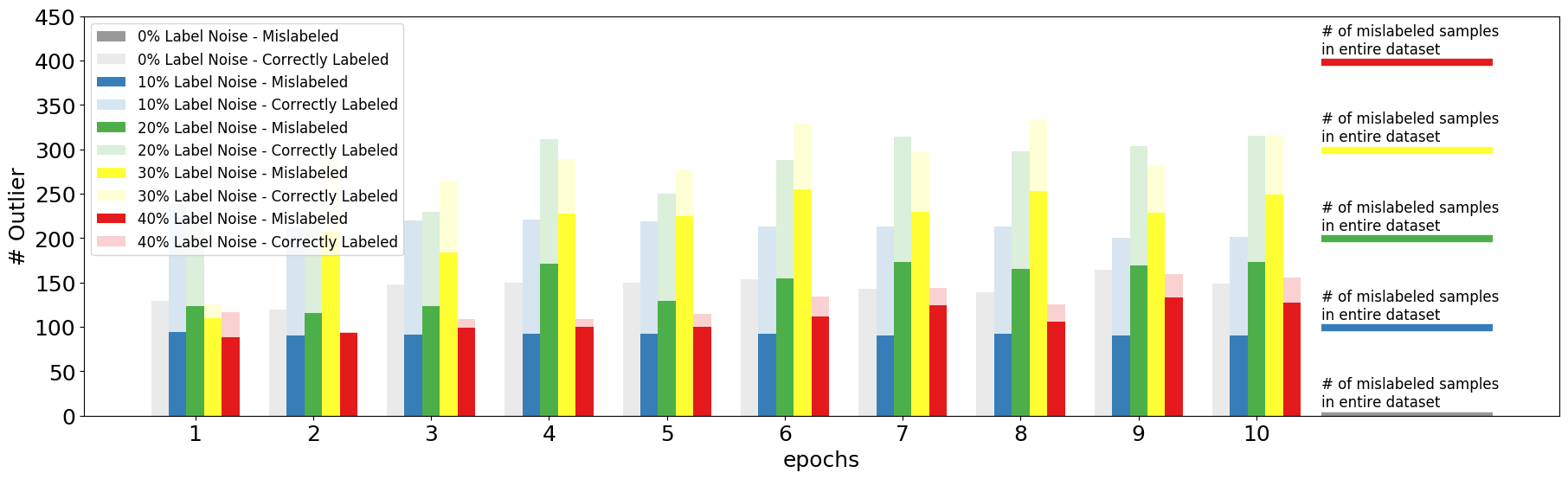}
   \caption{Building Defect Detection Dataset}
   \label{fig:OD_Build}
\end{subfigure}
\caption{No. of mislabeled and correctly labeled samples $\in \mathcal{O}$  for \textit{MixOutlier}.}
\end{figure*}

We evaluate the proposed algorithm for OD on the first 10 epochs on one iteration of the \textit{MixOutlier} approach, which is representative for all other approaches - see \figref{fig:OD_MNIST} and \figref{fig:OD_Build}, where mislabeled samples in the outlier dataset are distinguished from those that are correctly labeled.  

Given no label noise, only correctly labeled samples are identified as outliers. While the number of outliers quickly decreases on the MNIST dataset from 4521 (7.5\% of the entire dataset) to 281 (0.5\% of the entire dataset) within the first 10 epochs, it stays rather constant over the first 273 epochs with $141 \pm 18$ outlier samples on the BDD dataset. In the subsequent epochs, considerably fewer outliers are detected ($64 \pm 11$). 

Given medium noise levels, 
the detection of mislabeled samples is very efficient. 99\% of the mislabeled samples are identified within the first few epochs on the MNIST dataset for both noise settings. On the BDD dataset, 95\% of mislabeled samples are initially identified on the dataset with 10\% label noise and 87\% on the dataset with 20\% label noise. 
Yet the detection approach is lacking precision at the medium noise level as the set of outliers also includes correctly labeled samples. For example, on the BDD dataset  144 (on the dataset with 10\% label noise) and 141 (on the dataset with 20\% label noise) correctly labeled samples are initially in the set of outliers. This corresponds to 16\% and 18\% of all correctly labeled samples in the respective datasets. Yet, as the training continues, the number of mislabeled samples in the set of outliers stays approximately constant over all epochs ($88\% \pm 4\%$ and $81\% \pm 6\%$ of the truly mislabeled samples), whereas in the last training epoch, the number of correctly labeled samples decreases to 8\% and 9\% of the correctly labeled samples in the respective training datasets.

Given a massive noise level (40\% label noise ratio), the algorithm is more precise but less effective on both datasets. Initially on MNIST, 13\% of all mislabeled samples are detected and 2\% of all correctly labeled ones. As the training continues, up to 67\% of mislabeled samples are detected and 20\% of the correctly labeled ones.

\subsection{Experiment 4 - Binary Classification}
All introduced approaches combining the proposed OD with \textit{Adapted MixUp}, as introduced in \secref{sec:meth_combine}, are applied to the MNIST dataset and to the BDD dataset mislabeled by different ratios as described in \secref{sec:experiment}. The approaches are compared to ERM and \textit{mixup}. The results are shown in \tabref{table:results}. 

While all proposed approaches show similar performance to the baseline method \textit{mixup} on MNIST with 0 and 10\% label noise, they outperform \textit{mixup} on higher noise ratios. 
The performance gain compared to the baseline methods is most visible when using the \textit{MixAllOutlier+Relabel} approach on the MNIST dataset with a label noise ratio of 40\% (accuracy gain of 31\% accuracy compared to \textit{ERM} and 14\% compared to \textit{mixup}).

On the BDD dataset, the best-performing approach depends on the label noise ratio. On the datasets with little to medium noise levels, \textit{mixup} performs about as well as \textit{MixOutlier}, whereas all of the other proposed approaches perform worse. At massive noise levels, most of the proposed approaches outperform both baseline methods (\textit{ERM} and \textit{mixup}). Again, the model trained with \textit{MixAllOutlier+Relabel} on 40\% label noise achieves the biggest gain in accuracy compared to the models trained on the baseline methods. 

\ifdefined\SMC
\begin{table*}
\caption{Final Accuracy on Clean Test Dataset Trained on Mislabeled Datasets for Various Label Noise.}
\begin{center}
\begin{tabularx}{\textwidth}{l| *{5}{Y}|*{5}{Y}} 
\toprule
& \multicolumn{5}{c|}{\thead{\textbf{MNIST}}} & \multicolumn{5}{c}{\thead{\textbf{BDD}}}\\

\textbf{Method} \hfil - \hfil \textbf{Label Noise}  & \thead{\textbf{0\%}} &  \thead{\textbf{10\%}} &  \thead{\textbf{20\%}} &  \thead{\textbf{30\%}} &  \thead{\textbf{40\%}} & \thead{\textbf{0\%}} &  \thead{\textbf{10\%}} &  \thead{\textbf{20\%}} &  \thead{\textbf{30\%}} &  \thead{\textbf{40\%}}\\
\midrule
\midrule
ERM    &  $\mathbf{\SI{99 \pm 0}\%}$   & $\SI{91 \pm 1}\%$ & $\SI{82 \pm 1}\%$& $\SI{72 \pm 1}\%$ & $\SI{62 \pm 2}\%$ &  $\SI{94 \pm 1}\%$ & $\SI{90 \pm 2}\%$ & $\SI{83 \pm 1}\%$ & $\SI{79 \pm 1}\%$ & $\SI{63 \pm 4}\%$ \\

\textit{mixup}  &  $\SI{98 \pm 0}\%$  &  $\SI{97 \pm 0}\%$ &  $\SI{94 \pm 0}\%$ &  $\SI{91 \pm 1}\%$ &  $\SI{79 \pm 1}\%$ & $\SI{94 \pm 0}\%$ & $\mathbf{\SI{92 \pm 2}\%}$& $\mathbf{\SI{89 \pm 1}\%}$ & $\SI{83 \pm 2}\%$ & $\SI{73 \pm 5}\%$\\

DeleteOutlier    & $\SI{98 \pm 0}\%$  & $\mathbf{\SI{98 \pm 0}\%}$  & $\mathbf{\SI{97 \pm 0}\%}$  & $\SI{94 \pm 0}\%$ & $\SI{80 \pm 1}\%$   & $\SI{92 \pm 1}\%$ & $\SI{87 \pm 2}\%$ & $\SI{84 \pm 1}\%$ &$\mathbf{\SI{85 \pm 3}\%}$&$\SI{77 \pm 3}\%$\\

MixOutlier    & $\SI{98 \pm 0}\%$  &  $\mathbf{\SI{98 \pm 0}\%}$  &  $\mathbf{\SI{97 \pm 0}\%}$ &  $\SI{93 \pm 0}\%$  &  $\SI{80\pm 2}\%$   & $\mathbf{\SI{95 \pm 1}\%}$ & $\SI{91 \pm 1}\%$& $\mathbf{\SI{89 \pm 2}\%}$ &$\mathbf{\SI{85 \pm 3}\%}$ & $\SI{72 \pm 2}\%$   \\

MixAllOutlier  &   $\SI{98 \pm 0}\%$   & $\SI{97 \pm 0}\%$  & $\mathbf{\SI{97 \pm 0}\%}$  & $\SI{95 \pm 0}\%$  &$\SI{88\pm 1}\%$  &     $\SI{94 \pm 1}\%$& $\SI{85 \pm 3}\%$&   $\SI{87 \pm 3}\%$& $\SI{84 \pm 3}\%$  & $\SI{79 \pm 2}\%$   \\


MixAllOutlier+Relabel& $\SI{98 \pm 0}\%$    &  $\SI{97 \pm 0}\%$  &  $\mathbf{\SI{97 \pm 0}\%}$  &  $\mathbf{\SI{97 \pm 0}\%}$ &  $\mathbf{\SI{93 \pm 2}\%}$ & $\SI{94 \pm 1}\%$& $\SI{86 \pm 2}\%$& $\SI{86 \pm 2}\%$  & $\SI{84 \pm 2}\%$  & $\mathbf{\SI{83 \pm 3}\%}$ \\

\bottomrule
\end{tabularx}
\label{table:results}
\end{center}
\end{table*}
\fi

\ifdefined\ARXIV
\begin{table*}
\caption{Final Accuracy on Clean Test Dataset Trained on Mislabeled Datasets for Various Label Noise.}
\begin{center}
\begin{tabularx}{\textwidth}{l| *{5}{Y}|*{5}{Y}} 
\toprule
& \multicolumn{5}{c|}{\thead{\textbf{MNIST}}} & \multicolumn{5}{c}{\thead{\textbf{BDD}}}\\
\multicolumn{1}{r|}{\textbf{Label Noise}}  & \thead{\textbf{0\%}} &  \thead{\textbf{10\%}} &  \thead{\textbf{20\%}} &  \thead{\textbf{30\%}} &  \thead{\textbf{40\%}} & \thead{\textbf{0\%}} &  \thead{\textbf{10\%}} &  \thead{\textbf{20\%}} &  \thead{\textbf{30\%}} &  \thead{\textbf{40\%}}\\
 \textbf{Method}&&&&&&&&&\\
\midrule
\midrule
\small ERM    &  \small$\mathbf{\SI{99 \pm 0}\%}$   & \small$\SI{91 \pm 1}\%$ & \small$\SI{82 \pm 1}\%$& \small$\SI{72 \pm 1}\%$ & \small$\SI{62 \pm 2}\%$ &  \small$\SI{94 \pm 1}\%$ & \small$\SI{90 \pm 2}\%$ & \small$\SI{83 \pm 1}\%$ & \small$\SI{79 \pm 1}\%$ & \small$\SI{63 \pm 4}\%$ \\

\small \textit{mixup}  &  \small$\SI{98 \pm 0}\%$  &  \small$\SI{97 \pm 0}\%$ &  \small$\SI{94 \pm 0}\%$ &  \small$\SI{91 \pm 1}\%$ &  \small$\SI{79 \pm 1}\%$ & \small$\SI{94 \pm 0}\%$ & \small$\mathbf{\SI{92 \pm 2}\%}$& \small$\mathbf{\SI{89 \pm 1}\%}$ & \small$\SI{83 \pm 2}\%$ & \small$\SI{73 \pm 5}\%$\\

\small DeleteOutlier    & \small$\SI{98 \pm 0}\%$  & \small$\mathbf{\SI{98 \pm 0}\%}$  & \small$\mathbf{\SI{97 \pm 0}\%}$  & \small$\SI{94 \pm 0}\%$ & \small$\SI{80 \pm 1}\%$   & \small$\SI{92 \pm 1}\%$ & \small$\SI{87 \pm 2}\%$ & \small$\SI{84 \pm 1}\%$ &\small$\mathbf{\SI{85 \pm 3}\%}$&\small$\SI{77 \pm 3}\%$\\

\small MixOutlier    & \small$\SI{98 \pm 0}\%$  &  \small$\mathbf{\SI{98 \pm 0}\%}$  &  \small$\mathbf{\SI{97 \pm 0}\%}$ &  \small$\SI{93 \pm 0}\%$  &  \small$\SI{80\pm 2}\%$   & \small$\mathbf{\SI{95 \pm 1}\%}$ & \small$\SI{91 \pm 1}\%$& \small$\mathbf{\SI{89 \pm 2}\%}$ & \small$\mathbf{\SI{85 \pm 3}\%}$ & \small$\SI{72 \pm 2}\%$   \\

\small MixAllOutlier  &   \small$\SI{98 \pm 0}\%$   & \small$\SI{97 \pm 0}\%$  & \small$\mathbf{\SI{97 \pm 0}\%}$  & \small$\SI{95 \pm 0}\%$  & \small$\SI{88\pm 1}\%$  &     \small$\SI{94 \pm 1}\%$& \small$\SI{85 \pm 3}\%$&   \small$\SI{87 \pm 3}\%$& \small$\SI{84 \pm 3}\%$  & \small$\SI{79 \pm 2}\%$   \\


\multirow{2}{*}{\shortstack[c]{\small MixAllOutlier\\ \small +Relabel}}& \multirow{2}{*}{\small $\SI{98 \pm 0}\%$}    &  \multirow{2}{*}{\small$\SI{97 \pm 0}\%$}  &  \multirow{2}{*}{\small$\mathbf{\SI{97 \pm 0}\%}$}  &  \multirow{2}{*}{\small$\mathbf{\SI{97 \pm 0}\%}$} &  \multirow{2}{*}{\small$\mathbf{\SI{93 \pm 2}\%}$} & \multirow{2}{*}{\small$\SI{94 \pm 1}\%$}& \multirow{2}{*}{\small$\SI{86 \pm 2}\%$}& \multirow{2}{*}{\small$\SI{86 \pm 2}\%$}  & \multirow{2}{*}{\small$\SI{84 \pm 2}\%$}  & \multirow{2}{*}{\small$\mathbf{\SI{83 \pm 3}\%}$} \\
&&&&&&&&&\\
\bottomrule
\end{tabularx}
\label{table:results}
\end{center}
\end{table*}
\fi

\section{Discussion}
\label{sec:discussion}
\textbf{Training Dynamics on Two Datasets:} The models trained on both datasets provide a suitable testbed for evaluating the universality and sensitivity of the proposed approaches as they show different dynamics under label noise. 

\textbf{Outlier Detection:} Given the ground truth information about all labels, \textit{Adapted MixUp} augmentation  leads to a good generalization capability for both datasets and all label noise settings. However, the detection of mislabeled samples is a challenging task. While the  OD  shows similar behaviour on both datasets, its precision and effectiveness depends on the label noise ratio: 
it is more effective and less precise at medium noise levels and vice versa - more precise but less effective at massive noise levels. 
The lower precision at medium noise levels does not negatively impact the performance of the models trained on the larger MNIST dataset. 
It does, however, decrease the model's final performance on the smaller BDD dataset compared to the baseline method \textit{mixup} (e.g.\ with the \textit{DeleteOutlier}, \textit{MixAllOutlier}, or \textit{MixAllOutlier+Relabel} approach). The low effectiveness at massive noise levels especially affects the performance of models that are able to "memorize" the mislabeled samples faster than they are detected. This becomes particularly evident looking at the final performance of the models trained with  \textit{DeleteOutlier} or \textit{MixOutlier} on MNIST with 40\% label noise. 

\textbf{Generalization Capabilities of the Trained Models:} Contrary to the previous study of \citeauthor{zhang2017mixup} \cite{zhang2017mixup}, \textit{mixup} does not outperform \textit{ERM} in our experiments if no label noise (0\%) is present. This is most likely due to the fact that $\alpha$ has not been tuned but rather set based on assumptions about the upper limit of the label noise as defined in \secref{sec:Hyperparameter}. \textit{MixOutlier} slightly outperforms \textit{ERM} on the clean BDD training dataset. This might hint towards findings in the literature on curriculum learning where a curriculum that sorts the training dataset  can guide optimization towards a preferable optimum \cite{hacohen2019power}. However, the performance gain in our experiments is not significant and the results on MNIST do not support the hypothesis that the proposed approach acts as a curriculum that is beneficial for optimization. Therefore, this is left for future research. 

If the dataset is truly mislabeled (label noise ratio > 0\%), all of our proposed approaches along with  \textit{mixup}  outperform the baseline method \textit{ERM}. \textit{mixup} results in a satisfactory performance at medium noise levels on both datasets. However, \textit{MixOutlier} yields a superior performance on MNIST and a comparable performance on the BDD dataset. Most of the other proposed approaches in \secref{sec:meth_combine} suffer from the insufficient precision of the OD on the smaller BDD dataset, as  described above at medium noise levels.  
In the case of massive noise, the performance of the baseline method \textit{mixup} drops compared to the other noise levels. While some of the proposed approaches suffer from the initially low effectiveness of the OD (as decribed above), they still perform as well as or better than baseline method \textit{mixup}. Moreover, \textit{MixAllOutlier} and \textit{MixAllOutlier+Relabel} outperform \textit{mixup} significantly on both datasets at high noise levels (accuracy gain of 6-14\%). 

Based on the above evaluations, we recommend choosing the optimization methodology based on the assumption regarding the label noise. We recommend using \textit{MixOutlier} or the baseline \textit{mixup} for scenarios in which little to medium noise is suspected and the \textit{MixAllOutlier} approach for massive noise. 

\textbf{Sensitivity Analysis of the Recommendations:} To evaluate the sensitivity of these recommendations, the behaviour of the proposed approaches is evaluated for cases in which the assumption regarding the upper limit of label noise is incorrect. We assess only the best-performing approaches per noise level, i.e.\ \textbf{MixAllOutlier} and \textbf{MixOutlier+Relabel} (\tabref{table:sens_over}) for a massive assumed noise level, and \textbf{MixOutlier} (\tabref{table:sens_under}) for a little to medium assumed noise level. The performance is compared to the baseline method \textit{mixup} given the same assumptions, i.e.\ the HP setting of $\alpha$ as described in \tabref{table:MixUP_ParaMeter}. 

\begin{table}
\caption{Final Accuracy with Overestimated Noise Level.}
\begin{center}
\begin{tabularx}{\columnwidth}{l|l| l| *{3}{Y}} 
\toprule
 & \multirow{2}{*}{\makecell[bc]{\textbf{Assumed}\\\textbf{Noise} \\ \textbf{Limit}}} &  & \multicolumn{3}{c}{\thead{\textbf{Actual Label Noise}}} \\ 
 {\thead{\textbf{Dataset}}} & & \thead{\textbf{Method}} &    \thead{\textbf{0\%}} &  \thead{\textbf{10\%}} &  \thead{\textbf{20\%}} \\
\midrule
\multirow{4}{*}{MNIST} &\multirow{4}{*}{\shortstack[c]{\textit{Massive} \\ 50\%}} & \textit{mixup}  & $\mathbf{\SI{98 \pm 0}\%}$ & $\mathbf{\SI{97 \pm 0}\%}$ & $\SI{95 \pm 0}\%$\\
&&MixAllOutlier  & $\mathbf{\SI{98 \pm 0}\%}$ &  $\mathbf{\SI{97 \pm 0}\%}$ &  $\SI{96 \pm 0}\%$\\
&&\multirow{2}{*}{\shortstack[c]{MixAllOutlier\\+Relabel}}  & \multirow{2}{*}{$\SI{97 \pm 0}\%$} & \multirow{2}{*}{$\mathbf{\SI{97 \pm 0}\%}$} & \multirow{2}{*}{$\mathbf{\SI{97 \pm 0}\%}$} \\
&&&&&\\
\midrule
\multirow{4}{*}{BDD} &\multirow{4}{*}{\shortstack[c]{\textit{Massive} \\ 50\%}} & \textit{mixup}  & $\mathbf{\SI{94 \pm 0}\%}$ & $\mathbf{\SI{93 \pm 0}\%}$ & $\mathbf{\SI{89 \pm 2}\%}$\\
&&MixAllOutlier  &$\SI{92 \pm 1}\%$ &$\SI{89 \pm 2}\%$  &$\SI{87 \pm 2}\%$\\
&&\multirow{2}{*}{\shortstack[c]{MixAllOutlier\\+Relabel}}  & \multirow{2}{*}{$\SI{86 \pm 0}\%$} & \multirow{2}{*}{$\SI{85 \pm 1}\%$} & \multirow{2}{*}{$\SI{83 \pm 2}\%$} \\
&&&&&\\
\bottomrule
\end{tabularx}
\label{table:sens_over}
\end{center}
\end{table}

\begin{table}
\caption{Final Accuracy with Underestimated Noise Level.}
\begin{center}
\ifdefined\ARXIV
\begin{tabularx}{\linewidth}{l| c| l| *{3}{Y}}
\ifdefined\SMC
\begin{tabularx}{\linewidth}{l| l| l| *{3}{Y}}
\fi
\toprule
 & \multirow{2}{*}{\makecell[tc]{\textbf{Assumed}\\\textbf{Noise}\\\textbf{Limit}}} &  & \multicolumn{3}{c}{\thead{\textbf{Actual Label Noise}}} \\ 
 {\thead{\textbf{Dataset}}} & & \thead{\textbf{Method}} &    \thead{\textbf{20\%}} &  \thead{\textbf{30\%}} &  \thead{\textbf{40\%}} \\
\midrule
\multirow{4}{*}{MNIST} & \multirow{2}{*}{\shortstack[c]{\textit{Little} \\ 10\%}} & \textit{mixup}  & $\mathbf{\SI{89 \pm 1}\%}$ & $\mathbf{\SI{81 \pm 1}\%}$ & $\mathbf{\SI{69 \pm 1}\%}$\\
 & &MixOutlier  & $\SI{82 \pm 1}\%$ & $\SI{76 \pm 3}\%$ & $\mathbf{\SI{69 \pm 1}\%}$\\
\cmidrule{2-6}
 &\multirow{2}{*}{\shortstack[c]{\textit{Medium} \\ 30\%}} & \textit{mixup}  & $\SI{94 \pm 0}\%$ & $\SI{89 \pm 1}\%$ & $\SI{78 \pm 1}\%$\\
 & &MixOutlier  & $\mathbf{\SI{97 \pm 0}\%}$ & $\mathbf{\SI{93 \pm 2}\%}$ & $\mathbf{\SI{79 \pm 2}\%}$\\
\midrule
\multirow{4}{*}{BDD} & \multirow{2}{*}{\shortstack[c]{\textit{Little} \\ 10\%}} & \textit{mixup}  &  $\mathbf{\SI{90 \pm 2}\%}$ &  $\mathbf{\SI{82 \pm 2}\%}$&  $\mathbf{\SI{69 \pm 3}\%}$\\
 & & MixOutlier  &$\SI{86 \pm 2}\%$ & $\SI{80 \pm 2}\%$ &$\mathbf{\SI{69 \pm 2}\%}$\\
\cmidrule{2-6}
&\multirow{2}{*}{\shortstack[c]{\textit{Medium} \\ 30\%}} & \textit{mixup}  & $\mathbf{\SI{89 \pm 1}\%}$ &$\mathbf{\SI{86 \pm 1}\%}$ &$\mathbf{\SI{76 \pm 1}\%}$ \\
& & MixOutlier  & $\mathbf{\SI{89 \pm 2}\%}$ & $\SI{84 \pm 4}\%$ & $\SI{74 \pm 3}\%$\\
\bottomrule
\end{tabularx}
\end{center}
\label{table:sens_under}
\end{table}

The sensitivity analysis reveals that the baseline \textit{mixup} approach is less sensitive to erroneous assumptions of the label noise. This is particularly true if the noise level is underestimated, as the number of detected outliers surpasses the threshold and, therefore, hardly any outliers are considered.
However, if the noise level is overestimated, the best-performing approaches depend on the dataset. Yet the baseline \textit{mixup} shows, on average, the best performance over all settings and on both datasets. Hence, one drawback of the proposed framework is its sensitivity to inaccurate assumptions - especially if the noise is underestimated. However, these false assumptions can be easily identified: For example, if the number of detected outliers constantly surpasses the estimated upper noise level, it is a clear indication of faulty assumptions.

\section{Conclusion}
\label{sec:conclusion}
In this study, we proposed multiple variations of a two-step framework to train fault detection models that are robust to label noise. The framework first identifies outliers based on the hypothesis update and then modifies the training dataset accordingly. The framework's hyperparameters are set on the basis of a rough assumption regarding the label noise.  Ultimately, we demonstrate that the different strategies handle the level of noise and the uncertainty of this level very differently. Our proposed approaches outperform other approaches from the literature when the level of noise is expected to be high. In practical applications, it is considered realistic to obtain a good estimate regarding an upper noise limit. This is particularly the case for technical systems, where a rough assumption can be made as to how noisy the labels are expected to be (representing how unsure the domain experts are about the labels). In addition, our extensive evaluation can be used by practitioners to choose the appropriate approach based on a rough estimation of the upper limit of the label noise level. 
Lastly, our framework enables the early detection of outliers in the optimization process. This can provide additional information about the dataset that can be used for further evaluation. For example, analyzing the set of outliers already in the first training iteration provides information on the label noise in the dataset. Still, in real applications, the success of the optimization cannot be evaluated if no clean test dataset is available to measure the performance of the resulting model. However, the analysis of the detected outliers could be used to validate the model instead.
Evaluating the proposed methodology on more datasets is left for future research, particularly in terms of assessing the robustness of the approaches with respect to intra-class variability. Furthermore, evaluating the proposed OD in terms of its ability to act as a curriculum for efficient learning will be subject to further research.

\ifdefined\ARXIV
    \section*{Acknowledgment}
    \aknow
\fi
\printbibliography

@inproceedings{tanaka2018joint,
  title={Joint optimization framework for learning with noisy labels},
  author={Tanaka, Daiki and Ikami, Daiki and Yamasaki, Toshihiko and Aizawa, Kiyoharu},
  booktitle={Proceedings of the IEEE Conference on Computer Vision and Pattern Recognition},
  pages={5552--5560},
  year={2018}
}

@article{reed2014training,
  title={Training deep neural networks on noisy labels with bootstrapping},
  author={Reed, Scott and Lee, Honglak and Anguelov, Dragomir and Szegedy, Christian and Erhan, Dumitru and Rabinovich, Andrew},
  journal={CoRR, abs/1412.6596,},
  year={2014}
}

@inproceedings{ma2018dimensionality,
  title={Dimensionality-Driven Learning with Noisy Labels},
  author={Ma, Xingjun and Wang, Yisen and Houle, Michael E and Zhou, Shuo and Erfani, Sarah and Xia, Shutao and Wijewickrema, Sudanthi and Bailey, James},
  booktitle={International Conference on Machine Learning},
  pages={3355--3364},
  year={2018}
}

@inproceedings{hacohen2019power,
  title={On The Power of Curriculum Learning in Training Deep Networks},
  author={Hacohen, Guy and Weinshall, Daphna},
  booktitle={International Conference on Machine Learning},
  pages={2535--2544},
  year={2019}
}

@book{vapnik2013nature,
  author={Vapnik, Vladimir},
  year={2013},
  publisher={Springer science \& business media},
  title={The nature of statistical learning theory}
}

@inproceedings{chen2019understanding,
  title={Understanding and Utilizing Deep Neural Networks Trained with Noisy Labels},
  author={Chen, Pengfei and Liao, Ben Ben and Chen, Guangyong and Zhang, Shengyu},
  booktitle={International Conference on Machine Learning},
  pages={1062--1070},
  year={2019}
}

@article{zhang2016understanding,
  title={Understanding deep learning requires rethinking generalization},
  author={Zhang, Chiyuan and Bengio, Samy and Hardt, Moritz and Recht, Benjamin and Vinyals, Oriol},
  journal={International Conference on Learning Representations},
  year={2017}
}

@article{zhang2017mixup,
  title={mixup: Beyond empirical risk minimization},
  author={Zhang, Hongyi and Cisse, Moustapha and Dauphin, Yann N and Lopez-Paz, David},
  journal={arXiv preprint arXiv:1710.09412},
  year={2017}
}

@inproceedings{chapelle2001vicinal,
  title={Vicinal risk minimization},
  author={Chapelle, Olivier and Weston, Jason and Bottou, L{\'e}on and Vapnik, Vladimir},
  booktitle={Advances in neural information processing systems},
  pages={416--422},
  year={2001}
}

@article{vapnik1998statistical,
  title={Statistical learning theory.},
  author={Vapnik, Vladimir},
  journal={Wiley. Wang, K., Tsung, F.(2007). Run-to-run Process Adjust. using Categ. Obs. J. Qual. Technol.},
  volume={39},
  number={4},
  pages={312},
  year={1998}
}

@inproceedings{lecun1990handwritten,
  title={Handwritten digit recognition with a back-propagation network},
  author={LeCun, Yann and Boser, Bernhard E and Denker, John S and Henderson, Donnie and Howard, Richard E and Hubbard, Wayne E and Jackel, Lawrence D},
  booktitle={Advances in neural information processing systems},
  pages={396--404},
  year={1990}
}

@article{abdeljaber2017real,
  title={Real-time vibration-based structural damage detection using one-dimensional convolutional neural networks},
  author={Abdeljaber, Osama and Avci, Onur and Kiranyaz, Serkan and Gabbouj, Moncef and Inman, Daniel J},
  journal={Journal of Sound and Vibration},
  volume={388},
  pages={154--170},
  year={2017},
  publisher={Elsevier}
}

@article{krummenacher2017wheel,
  title={Wheel defect detection with machine learning},
  author={Krummenacher, Gabriel and Ong, Cheng Soon and Koller, Stefan and Kobayashi, Seijin and Buhmann, Joachim M},
  journal={IEEE Transactions on Intelligent Transportation Systems},
  volume={19},
  number={4},
  pages={1176--1187},
  year={2017},
  publisher={IEEE}
}

@inproceedings{muller2019identifying,
  title={Identifying Mislabeled Instances in Classification Datasets},
  author={M{\"u}ller, Nicolas M and Markert, Karla},
  booktitle={2019 International Joint Conference on Neural Networks},
  pages={1--8},
  year={2019},
  organization={IEEE}
}

@article{jiang2017mentornet,
  title={Mentornet: Learning data-driven curriculum for very deep neural networks on corrupted labels},
  author={Jiang, Lu and Zhou, Zhengyuan and Leung, Thomas and Li, Li-Jia and Fei-Fei, Li},
  journal={CoRR, abs/1712.05055},
  year={2017}
}

@inproceedings{arpit2017closer,
  title={A closer look at memorization in deep networks},
  author={Arpit, Devansh and Jastrz{\k{e}}bski, Stanis{\l}aw and Ballas, Nicolas and Krueger, David and Bengio, Emmanuel and Kanwal, Maxinder S and Maharaj, Tegan and Fischer, Asja and Courville, Aaron and Bengio, Yoshua and others},
  booktitle={Proceedings of the 34th International Conference on Machine Learning-Volume 70},
  pages={233--242},
  year={2017}
}

@article{frenay2013classification,
  title={Classification in the presence of label noise: a survey},
  author={Fr{\'e}nay, Beno{\^\i}t and Verleysen, Michel},
  journal={IEEE transactions on neural networks and learning systems},
  volume={25},
  number={5},
  pages={845--869},
  year={2013},
  publisher={IEEE}
}

@article{pleiss2020identifying,
  title={Identifying Mislabeled Data using the Area Under the Margin Ranking},
  author={Pleiss, Geoff and Zhang, Tianyi and Elenberg, Ethan R and Weinberger, Kilian Q},
  journal={arXiv preprint arXiv:2001.10528},
  year={2020}
}

@inproceedings{patrini2017making,
  title={Making deep neural networks robust to label noise: A loss correction approach},
  author={Patrini, Giorgio and Rozza, Alessandro and Krishna Menon, Aditya and Nock, Richard and Qu, Lizhen},
  booktitle={Proceedings of the IEEE Conference on Computer Vision and Pattern Recognition},
  pages={1944--1952},
  year={2017}
}

@inproceedings{hu2020simple,
  title={Simple and effective regularization methods for training on noisily labeled data with generalization guarantee},
  author={Hu, W and Li, Z and Yu, D},
  booktitle={International Conference on Learning Representations},
  year={2020}
}

@article{shen2018learning,
  title={Learning with bad training data via iterative trimmed loss minimization},
  author={Shen, Yanyao and Sanghavi, Sujay},
  journal={International Conference on Machine Learning},
  year={2018}
}

@article{ren2018learning,
  title={Learning to reweight examples for robust deep learning},
  author={Ren, Mengye and Zeng, Wenyuan and Yang, Bin and Urtasun, Raquel},
  journal={International Conference on Machine Learning},
  year={2018}
}

@inproceedings{vahdat2017toward,
  title={Toward robustness against label noise in training deep discriminative neural networks},
  author={Vahdat, Arash},
  booktitle={Advances in Neural Information Processing Systems},
  pages={5596--5605},
  year={2017}
}

@article{kingma2014adam,
  title={Adam: A method for stochastic optimization},
  author={Kingma, Diederik P and Ba, Jimmy},
  journal={arXiv preprint arXiv:1412.6980},
  year={2014}
}

@article{sukhbaatar2014training,
  title={Training convolutional networks with noisy labels},
  author={Sukhbaatar, Sainbayar and Bruna, Joan and Paluri, Manohar and Bourdev, Lubomir and Fergus, Rob},
  journal={Proc. Int. Conf. Learn. Represent. Workshop, 2015},
  pages={1–11},
  year={2015}
}

@inproceedings{xu2019l_dmi,
  title={{$\mathcal{L}_{DMI}$}: A Novel Information-theoretic Loss Function for Training Deep Nets Robust to Label Noise},
  author={Xu, Yilun and Cao, Peng and Kong, Yuqing and Wang, Yizhou},
  booktitle={Advances in Neural Information Processing Systems},
  pages={6222--6233},
  year={2019}
}

@inproceedings{zhang2018generalized,
  title={Generalized cross entropy loss for training deep neural networks with noisy labels},
  author={Zhang, Zhilu and Sabuncu, Mert},
  booktitle={Advances in neural information processing systems},
  pages={8778--8788},
  year={2018}
}

@article{ducoffeanomaly,
  title={Anomaly Detection on Time Series with Wasserstein GAN applied to PHM},
  author={Ducoffe, M{\'e}lanie and Haloui, Ilyass and Gupta, Jayant Sen and Supaero, ISAE},
  journal = {International Journal of Prognostics and Health Management},
  year={2019}
}

@inproceedings{krishna2016embracing,
  title={Embracing error to enable rapid crowdsourcing},
  author={Krishna, Ranjay A and Hata, Kenji and Chen, Stephanie and Kravitz, Joshua and Shamma, David A and Fei-Fei, Li and Bernstein, Michael S},
  booktitle={Proceedings of the 2016 CHI conference on human factors in computing systems},
  pages={3167--3179},
  year={2016}
}

@article{granderson2020building,
  title={Building fault detection data to aid diagnostic algorithm creation and performance testing},
  author={Granderson, Jessica and Lin, Guanjing and Harding, Ari and Im, Piljae and Chen, Yan},
  journal={Scientific Data},
  volume={7},
  number={1},
  pages={1--14},
  year={2020},
  publisher={Nature Publishing Group}
}

\end{document}